\documentclass{article}
\usepackage{ijcai24}

\usepackage{times}
\usepackage{soul}
\usepackage{url}
\usepackage[utf8]{inputenc}
\usepackage[small]{caption}
\usepackage{graphicx}
\usepackage{amsmath}
\usepackage{amsthm}
\usepackage{booktabs}
\usepackage{algorithm}
\usepackage{algorithmic}
\usepackage[switch]{lineno}

\usepackage{amssymb}
\usepackage{multirow}
\usepackage{colortbl}
\usepackage{xcolor}
\usepackage{array}
\usepackage{pifont}
\usepackage{bm}
\usepackage[misc]{ifsym}
\usepackage[pagebackref=true,breaklinks=true,letterpaper=true,colorlinks,bookmarks=false]{hyperref}

\def\eg{{\it{e.g.}}}

\definecolor{mycolor}{RGB}{241,240,255}

\title{Pre-training Point Cloud Compact Model with Partial-aware Reconstruction}

\author{%
  \  Yaohua Zha$^{1,2}$
  ~~ Yanzi Wang$^1$ 
  ~~ Tao Dai$^3$ \thanks{Corresponding author.} 
  ~~ Shu-Tao Xia$^{1,2}$ \vspace{0.3cm} \\
  \normalsize $^1$Tsinghua Shenzhen International Graduate School, Tsinghua University \\
  ~~ $^2$Research Center of Artificial Intelligence, Peng Cheng Laboratory\\
  ~~ $^3$College of Computer Science and Software Engineering, Shenzhen University\\
}

\begin{document}

\maketitle

\begin{abstract}
The pre-trained point cloud model based on Masked Point Modeling (MPM) has exhibited substantial improvements across various tasks. However, two drawbacks hinder their practical application. Firstly, the positional embedding of masked patches in the decoder results in the leakage of their central coordinates, leading to limited 3D representations. Secondly, the excessive model size of existing MPM methods results in higher demands for devices. To address these, we propose to pre-train \textbf{Point} cloud \textbf{C}ompact Model with \textbf{P}artial-aware \textbf{R}econstruction, named \textbf{Point-CPR}. Specifically, in the decoder, we couple the vanilla masked tokens with their positional embeddings as randomly masked queries and introduce a partial-aware prediction module before each decoder layer to predict them from the unmasked partial. It prevents the decoder from creating a shortcut between the central coordinates of masked patches and their reconstructed coordinates, enhancing the robustness of models. We also devise a compact encoder composed of local aggregation and MLPs, reducing the parameters and computational requirements compared to existing Transformer-based encoders. Extensive experiments demonstrate that our model exhibits strong performance across various tasks, especially surpassing the leading MPM-based model PointGPT-B with only \textbf{2\%} of its parameters. The code will be released.
\end{abstract}

\section{Introduction}

3D point cloud perception, as a crucial application of deep learning, has achieved significant success across various areas such as autonomous driving, robotics, and virtual reality. 
Point cloud self-supervised learning~\cite{pointcontrast,crosspoint,pointbert}, capable of learning universal representations from extensive unlabeled point cloud data, has gained much attention. Masked point modeling (MPM)~\cite{pointbert,pointmae,maskpoint,femae}, as an important self-supervised paradigm, has become mainstream in point cloud analysis and has gained immense success across diverse point cloud tasks.

The classical MPM~\cite{pointmae} is inspired by masked image modeling (MIM)~\cite{bao2021beit,mae,simmim} and begins by dividing the point cloud into patches, utilizing the central point coordinate of each patch to denote their positions. The patch content is represented by the relative coordinates of other points concerning the central point. By random masking and reconstruction, it predicts the relative coordinates of the masked patches. Despite the significant success, two drawbacks continue to limit the practical application of these MPM-based models.

\begin{figure}[t]
    \begin{center}
    \includegraphics[width=\linewidth]{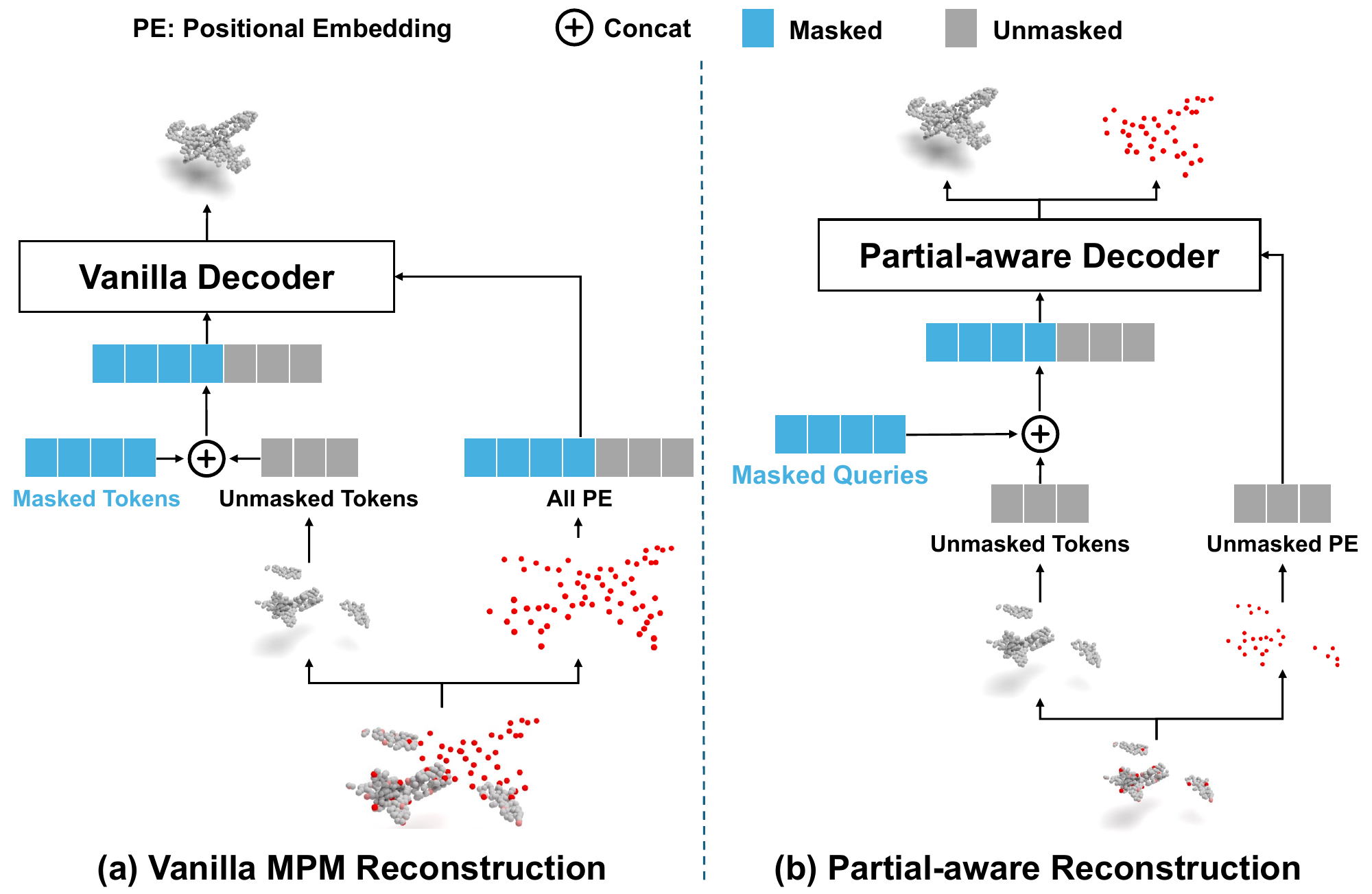}
    \caption{Comparison of (a) vanilla MPM reconstruction and (b) our partial-aware reconstruction. Our reconstruction does not require the center coordinates of masked patches as input. The encoding process is omitted in the figure.
    }\label{input}
    \end{center}
\end{figure}

The introduction of positional embedding of masked patches in the decoder results in the leakage of the central coordinates, leading to limited 3D representations. In classical MIM~\cite{mae}, the sequence order of each image patch provides positional embedding, while their content, \eg pixels, offers semantic details. Reconstructing the masked patches' semantic content relies on their positional embedding. However, in most point cloud data, the points are disordered, and only point coordinates are available. Existing MPM methods utilize the absolute coordinates of the central point of each patch to represent its position, while the relative coordinates of the other points in a patch to the central point represent its semantics, leading to a coupling of semantics and position. When reconstructing masked patches, the center point coordinates of masked patches are fed into the decoder as positional embeddings to predict their semantics, \eg relative coordinates. As depicted in Figure \ref{input} (a), the overall contour of the complete point cloud remains available during reconstruction. While this approach does alleviate learning difficulties caused by coordinates autoregression, existing methods that direct transfer from MIM may lead to shortcuts in learning reconstructed semantic coordinates due to the leakage of the central point coordinates of the patch, resulting in limited 3D representations.

\begin{figure}[t]
    \begin{center}
    \includegraphics[width=\linewidth]{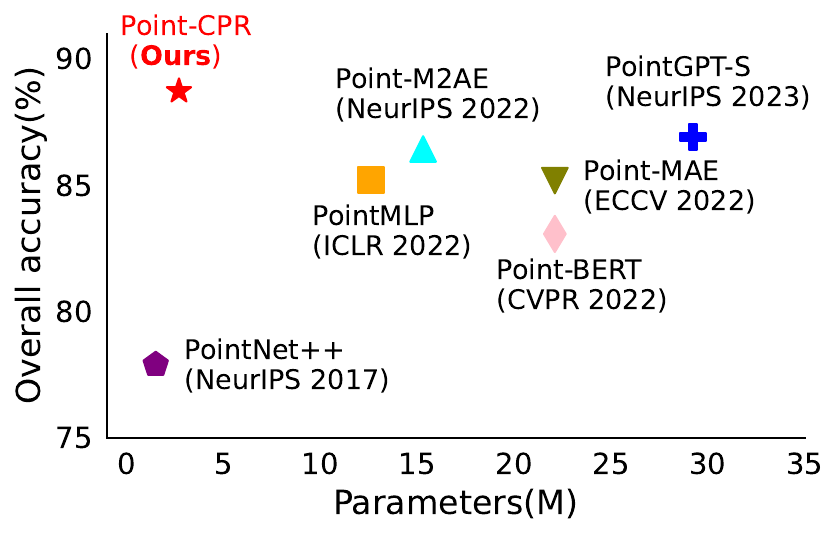}
    \caption{Accuracy-parameters tradeoff on ScanObjectNN. Our Point-CPR performs best. Please refer to Section \ref{subsec:exp} for details.
    }\label{params}
    \end{center}
\end{figure}

Another factor hindering the practical application of these models is their significant demands on both model size and computational complexity, imposing high requirements on practical devices. 
Indeed, in practical applications of point clouds, models are often deployed on embedded devices such as robots or VR headsets, where strict constraints exist regarding the model's size and complexity.
As shown in Fig \ref{params}, PointNet++~\cite{pointnet++}, the most popular point cloud analysis model, has merely 1.5M parameters for classification. In contrast, MPM methods like Point-MAE~\cite{pointmae} require 22.1M parameters and complexity exponentially grows with an increase in the number of points. This significant disparity imposes even higher demands on practical devices, particularly when facing limited resources.

To solve the above-mentioned issues, we propose to pre-train \textbf{Point} cloud \textbf{C}ompact Model with \textbf{P}artial-aware \textbf{R}econstruction, named \textbf{Point-CPR}. Specifically, unlike the existing approach of embedding each masked patch by combining positional embedding and its' mask tokens in the decoder, we assigned randomly initialized masked query to represent the combined token of each masked patch. At this point, as shown in Figure \ref{input}(b), the central coordinates of each masked patch are unavailable to the decoder. Therefore, we design a partial-aware prediction module before each layer of the decoder to predict tokens of the masked patches from the unmasked partial point cloud. 
Due to the decoded masked queries coupling the positional and semantic information, we not only reconstruct the semantic relative coordinates of the masked patches but also reconstruct their center coordinates. This dual reconstruction of semantics and positions disrupts the learning shortcuts caused by positional leakage in existing MPMs, making the pre-training more challenging and conducive to getting robust 3D representation.

To mitigate the high demands of pre-trained models on practical devices, we also devise a compact encoder to replace the vanilla Transformer-based encoder~\cite{attention}. Our compact encoder is exclusively composed of lightweight local aggregation modules and residual MLPs, thereby eschewing the computational complexity and continued memory access inherent in Self-Attention mechanisms. By doing so, as depicted in Figure \ref{params}, our pre-trained model exhibits a mere 2.7M parameters, significantly smaller than the 22.1M of Point-MAE and 29.2M of PointGPT-S~\cite{pointgpt}. Furthermore, these reduced parameters mitigate the risk of overfitting in downstream tasks, allowing the model's performance to surpass existing MPM models. 

Our main contributions are summarized as follows:

\begin{itemize}
    \item We propose a partial-aware reconstruction strategy for point cloud self-supervised learning, mitigating the issue of positional leakage during existing reconstruction caused by the inherent disorder and coordinate-based nature of point clouds.
    \item We design a compact encoder, that mitigates the high demands of pre-trained models on practical devices and also mitigates the risk of overfitting in downstream tasks.
    \item Extensive experiments demonstrate that our model exhibits strong performance across various tasks, especially surpassing the leading MPM-based model PointGPT-B with only \textbf{2\% }of its parameters.
\end{itemize}

\begin{figure*}[t]
    \begin{center}
    \includegraphics[width=\linewidth]{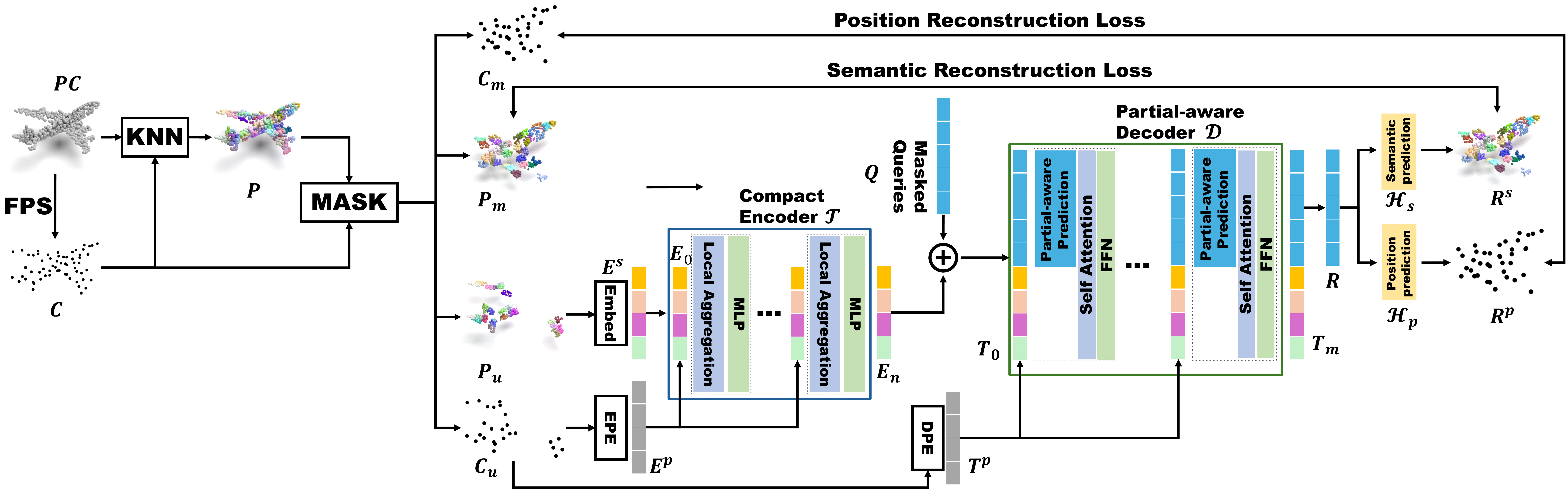}
    \caption{The pipeline of our Point-CPR. Given a point cloud, we first encode unmasked features by our compact encoder. Then, we concatenate random masked queries with the encoded features and feed them into our partial-aware decoder to decode the masked portion. Finally, we perform dual reconstruction of semantic and positional coordinates of masked patches.
    }\label{framework}
    \end{center}
\end{figure*}

\section{Related Work}

\subsection{Point Cloud Self-supervised Pre-training}
Point cloud self-supervised pre-training has achieved remarkable improvement in many point cloud tasks. This approach first applies a pretext task to learn the latent 3D representation and then transfers it to various downstream tasks. PointContrast~\cite{pointcontrast} and CrossPoint~\cite{crosspoint} initially explored utilizing contrastive learning~\cite{cpc,cmc} for learning 3D representations, which achieved some success; however, there were still some shortcomings in capturing fine-grained semantic representations. Recently, masked point modeling methods~\cite{pointbert,pointmae,maskpoint,pointgpt,femae,lcm} demonstrated significant improvements in learning fine-grained point cloud representations through masking and reconstruction. However, they still learned limited 3D representations. 
Many methods~\cite{act,jointmae,pimae} have attempted to leverage multimodal knowledge to assist MPM in learning more generalized representations, yielding significant improvements but also introducing additional computational pressure.
In this paper, we unleash the full potential of single-modal point cloud masking reconstruction by addressing the issue of positional leakage during reconstruction in existing MPMs, enabling the acquisition of more robust 3D representations.

\subsection{Deep network architecture for point cloud}
Point clouds, as 3D data directly sampled from scanning devices, inherently exhibit irregularity and disorder. To employ deep neural networks for point cloud analysis, various structures have been developed~\cite{pointnet,dgcnn,sfr,xiong2023semantic,pointmamba,dapt}. PointNet~\cite{pointnet}, a pioneer in point cloud analysis, introduced an MLP-based network to address the disorder of point clouds. Subsequently, PointNet++~\cite{pointnet++} further proposed adaptive aggregation of multiscale features on MLPs and incorporated local point sets for effective feature learning. DGCNN~\cite{dgcnn} introduced the graph convolutional networks dynamically computing local graph neighboring nodes to extract geometric information. PointMLP~\cite{pointmlp} suggested efficient point cloud representation solely relying on pure residual MLPs. Recently, many Transformer-based models~\cite{pct,3detr,pointmae,ptv2,ptv3}, benefiting from attention mechanisms, have achieved notable improvements in point cloud analysis. However, this led to a significant increase in model size, posing considerable challenges for practical applications. In this paper, we focus on designing more compact point cloud network architectures specific to pre-training models.

\section{Methodology}

The overall pipeline of our Point-CPR is shown in Figure \ref{framework}, primarily composed of a mask and embedding layer, a compact encoder, and a partial-aware decoder for reconstruction. In this section, we first introduce the overall pre-training pipeline of our Point-CPR (\S~\ref{subsec:pipe}). Next, we provide a detailed exposition of the design of the compact encoder employed for efficient feature extraction (\S~\ref{subsec:encoder}). Finally, we detail the design of our partial-aware decoder for robust 3D representations (\S~\ref{subsec:decoder}).

\subsection{The Pipeline of Point-CPR}
\label{subsec:pipe}

\subsubsection{Patching, Masking, and Embedding} 
Given an input point cloud $\bm {PC}\in \mathbb{R}^{N\times 3}$ with $N$ points, we initially downsample a central point cloud $\bm {C}\in \mathbb{R}^{M\times 3}$ with $M$ points by farthest point sampling (FPS). Then, we perform K-Nearest Neighborhood (KNN) around $\bm {C}$ to divide  $\bm {PC}$ into $M$ point patches $\bm {P}\in \mathbb{R}^{M \times K \times 3}$. Following this, we randomly mask a portion of $\bm {C}$ and $\bm {P}$, resulting in masked elements $\bm {C_m}\in \mathbb{R}^{(1-r)M\times 3}$ and $\bm {P_m}\in \mathbb{R}^{(1-r)M\times K \times 3}$, along with unmasked elements $\bm {C_u}\in \mathbb{R}^{rM\times 3}$ and $\bm {P_u}\in \mathbb{R}^{rM\times K \times 3}$, where $r$ denotes the unmask ratio. Finally, we use lightweight PointNet~\cite{pointnet} and MLP as semantic embedding layer (Embed) and position embedding layer (PE) respectively to extract semantic tokens $\bm {E^s}\in \mathbb{R}^{rM\times d}$ and central position embedding $\bm {E_p}\in \mathbb{R}^{rM\times d}$ for the unmasked patches. These embeddings are then added to obtain the initial features $\bm {E_0}\in \mathbb{R}^{rM\times d}$, where $d$ is the feature dimension.

\subsubsection{Encoder}
We employ our compact encoder ${\mathcal T}$ to extract features from the unmasked features $\bm {E_0}$. This encoder consists of a series of $ n$ encode layers, each layer incorporating a local aggregation module and an MLP layer, detailed in Figure \ref{encoder}. For the input feature $\bm {E_{i-1}}$ of the $i$-th layer, after adding its positional embedding $\bm {E^p}$, it feeds to the $i$-th encoding layer ${\mathcal T}_i$ to obtain the feature $\bm {E_i}$. Therefore, the forward process of each encoder layer is defined as:
\begin{gather}
    \label{eq1}
    \bm {E_i} = {\mathcal T}_i(\bm {E_{i-1}} + \bm {E^p}), \quad i = 1,...,n
\end{gather}

\subsubsection{Decoder}
In the decoding stage, unlike previous MPM methods that require fusing masked tokens and positional embeddings to represent the features of masked patches, we use randomly initialized mask queries $\bm Q\in \mathbb{R}^{rM\times d}$ to represent them. Subsequently, we employ our partial-aware decoder ${\mathcal D}$ to decode the masked features. Specifically, we first concatenate the features $\bm {E_{n}}\in \mathbb{R}^{rM\times d}$ of the unmasked patches with the randomly initialized mask queries $\bm Q$ to obtain the input $\bm {T_{0}}\in \mathbb{R}^{M\times d}$ for the decoder. Then, we embed the decoder positional embedding $\bm {T^{p}}\in \mathbb{R}^{rM\times d}$ of the unmasked patches. Finally, for the input feature $\bm {T_{i-1}}$ of the $i$-th decoder layer, we add the unmasked tokens (\eg the first $rM$ tokens $\bm {T_{i-1}}[:rM]$) to their positional embeddings $\bm {T^{p}}$ and concatenate the results with the next $rM$ masked query tokens $\bm {T_{i-1}}[rM:]$. Therefore, the forward process of each decoder layer is defined as:
\begin{gather}
    \label{eq2}
    \bm {{T}_i} = {\mathcal D}_i({\bm \{\bm {T_{i-1}}[:rM] + \bm {T^p} \bm ; \bm {T_{i-1}}[rM:]\bm \}_{\bm 0}}), i = 1,...,m,
\end{gather}
where ${\bm {\{;\}}_0}$ denotes concatenation along the token dimension.

\subsubsection{Reconstruction}
We utilize the features $\bm R =\bm {T_m}[rM:]$ decoded by the decoder to perform the 3D reconstruction. In contrast to previous MPM methods that only reconstruct semantics, we reconstruct the masked patch from both semantic and central coordinates. We employ multi-layer MLPs to construct semantic reconstruction head ${\mathcal H_s}$ and central position reconstruction head ${\mathcal H_p}$ and our reconstruction target is to recover the central coordinates $\bm R_{p} = \mathcal H_p(\bm R)$ and relative coordinates $\bm R_{s} = \mathcal H_s(\bm R)$ of the masked patches. 

We use the $\bm {l_2}$ Chamfer Distance \cite{cdloss} ($\mathcal {CD}$) as our reconstruction loss. Therefore, our loss function $\mathcal {L}$ is as follows
\begin{gather}
    \label{eq1}
    \mathcal {L}=\mathcal {CD}(\bm {R^s},\bm {P_m})+ \mathcal {CD}(\bm {R^p},\bm {C_m})
\end{gather}

\begin{figure}[t]
    \begin{center}
    \includegraphics[width=\linewidth]{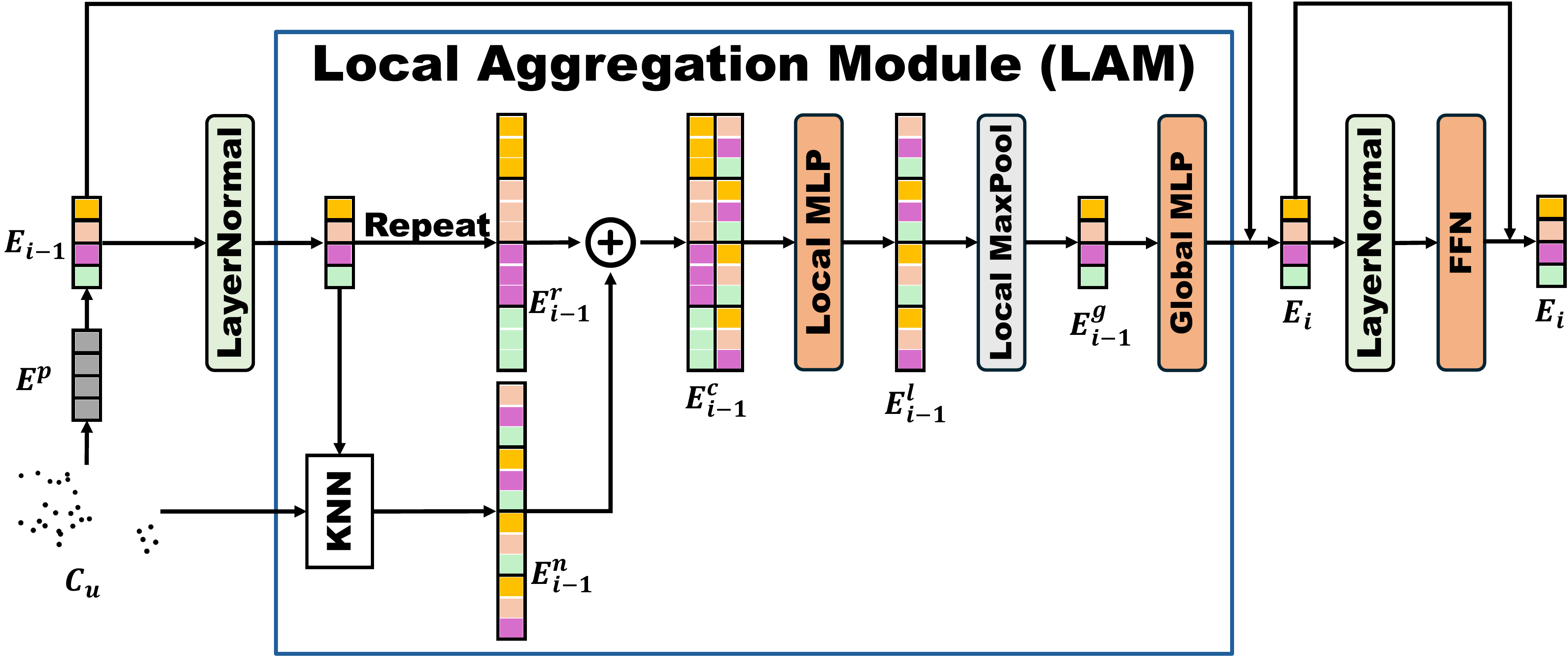}
    \caption{The structure of our compact encoder layer and our compact encoder consists of $n$ stacked compact encoder layers.
    }\label{encoder}
    \end{center}
\end{figure}
\subsection{Compact Encoder}
\label{subsec:encoder}

The classical Transformer~\cite{attention} relies on the Self-Attention mechanism to perceive long-range correlations among all patches globally and has achieved great success in language and image domains. However, there remains uncertainty about whether directly transferring a Transformer-based encoder is suitable for point cloud data. Firstly, applications of point clouds are more inclined towards practical consumer devices, such as VR glasses and automobiles. The hardware resources of these devices are limited, imposing higher limits on the model size and complexity, and the Transformer-based backbone demands significantly more resources than traditional networks, as illustrated in Figure \ref{params}. Secondly, extensive research~\cite{pointnet++,dgcnn,3detr} also indicates that the perception of local geometry in point cloud data far outweighs the need for global perception. Therefore, the computation of long-range correlations in Self-Attention leads to a considerable amount of redundant calculations. To address these practical issues, we propose a compact encoder based on local aggregation.

Our compact encoder consists of $n$ stacked compact encoder layers, each layer comprising a local aggregation module (LAM) and a feed-forward network (FFN), as shown in Figure \ref{encoder}. 
For the $i$-th encoder layer, the output ($\bm {E_{i-1}}$) of the preceding layer, augmented with the positional embedding and layer normal, is initially fed to the Local Aggregation Module (LAM) for aggregating local geometric. Afterward, the result is added to the input residual, passed through layer normalization, and finally fed into a Feed-forward Network (FFN) to obtain the ultimate output feature ($\bm {E_{i}}$). This process can be formalized as follows:
\begin{gather}
    \label{eq1}
    \bm {E_{i}} = \bm {E_{i-1}} + l_i(n_i^1(\bm {E_{i-1}}), \bm {C_{u}}) \\
    \bm {E_{i}} = \bm {E_{i}} + f_i(n_i^2(\bm {E_{i}}))
\end{gather}
where $l(\cdot)$ represents the LAM, $n(\cdot)$ represents layer normalization, and $f(\cdot)$ represents the FFN.

In the Local Aggregation Module, we first use the k-nearest neighbors algorithm based on the features $\bm {E_{i-1}}$ and its central coordinates $\bm {C_{u}}$ to find the $k$ nearest neighbors feature $\bm {E_{i-1}^n}\in \mathbb{R}^{(1-r)kM\times C}$ for each token in $\bm {E_{i-1}}$. We then replicate each token of $\bm {E_{i-1}}$ $k$ times and concatenate them with their corresponding neighbors to obtain $\bm {E_{i-1}^c}\in \mathbb{R}^{(1-r)M\times 2C}$. 
Next, Local MLP performs a non-linear mapping on all local neighboring features to capture local geometric information. Subsequently, local max pooling is applied to aggregate all local features for each patch. Finally, Global MLP maps all patches to obtain locally enhanced features $\bm {E_{i}}\in \mathbb{R}^{(1-r)M\times C}$. Our LAM consists of only two simple MLP layers, significantly reducing computational requirements compared to redundant Self-Attention mechanisms. Additionally, further experimental analysis indicates that our compact encoder, due to a significant reduction in parameters, can effectively alleviate the overfitting issues associated with Transformers in downstream tasks. Please refer to supplementary files for details.

% Table generated by Excel2LaTeX from sheet 'Sheet5'
\begin{table*}[t]
  \centering
  \resizebox{\textwidth}{!}{
    \begin{tabular}{lccccccccc}
    \toprule
    \multirow{2}[4]{*}{Method} & \multirow{2}[4]{*}{Reference} & \multirow{2}[4]{*}{Pre-training} & \multicolumn{1}{l}{\multirow{2}[4]{*}{\#Params (M)}} & \multirow{2}[4]{*}{GFLOPs} & \multicolumn{3}{c}{ScanObjectNN} & \multicolumn{2}{c}{ModelNet40} \\
    \cmidrule(lr){6-8}\cmidrule(lr){9-10}\textbf{}          &       &       &       &       & OBJ-BG & OBJ-ONLY & PB-T50-RS & w/o Vote & w/ Vote \\
    \midrule
    \multicolumn{10}{c}{\textit{Supervised Learning Only}} \\
    \midrule
    PointNe~\cite{pointnet} & CVPR 2017  & \ding{56}     & 3.5   & 0.5   & 73.3  & 79.2  & 68    & 89.2  & - \\
    PointNet++~\cite{pointnet++} & NeurIPS 2017 & \ding{56}     & 1.5   & 1.7   & 82.3  & 84.3  & 77.9  & 90.7  & - \\
    DGCNN~\cite{dgcnn} & TOG 2019 & \ding{56}     & 1.8   & 2.4   & 82.8  & 86.2  & 78.1  & 92.9  & - \\
    MVTN~\cite{MVTN}  & ICCV 2021 & \ding{56}     &  11.2 & 43.7  &  92.6 &  92.3 & 82.8  & 93.8  & - \\
    PointMLP~\cite{pointmlp} & ICLR 2022 & \ding{56}     & 12.6  & 31.4  & -     & -     & 85.2  & 94.1  & 94.5 \\
    P2P-HorNet~\cite{p2p} & NeurIPS 2022 & \ding{56}     & 195.8 & 34.6  & -     & -     & 89.3  & 94.0  & - \\
    \midrule
    \multicolumn{10}{c}{\textit{Single Modal Self-Supervised Learning}} \\
    \midrule
    STRL~\cite{strl}  & ICCV 2021 & CL    & -     & -     & -     & -     & -     & 93.1  & - \\
    Point-BERT~\cite{pointbert} & CVPR 2022 & MPM   & 22.1  & 4.8   & 87.43 & 88.12 & 83.07 & 92.7  & 93.2 \\
    % MaskPoint~\cite{maskpoint} & ECCV 2022 & MPM   & 22.1  & 4.8   & 89.30 & 88.10 & 84.30 & -     & 93.8 \\
    Point-MAE~\cite{pointmae} & ECCV 2022 & MPM   & 22.1  & 4.8   & 90.02 & 88.29 & 85.18 & 93.2  & 93.8 \\
    Point-M2AE~\cite{m2ae} & NeurIPS 2022 & MPM   & 15.3  & 3.6   & 91.22 & 88.81 & 86.43 & 93.4  & 94.0 \\
    PointGPT-S~\cite{pointgpt} & NeurIPS 2023 & MPM   & 29.2  & 4.5   & 91.63 & 90.02 & 86.88 & - & 94.0 \\
    PointGPT-B~\cite{pointgpt} & NeurIPS 2023 & MPM   & 120.5 & 36.2  & 93.60 & 92.50 & \textbf{89.60} & - & 94.2 \\
    IDPT~\cite{idpt} & ICCV 2023 & MPM   & 23.5 & -  & 93.63 & 93.12 & 88.51 & 93.3 & 94.4 \\
    PointFEMAE~\cite{femae} & AAAI 2024 & MPM   & 27.4 & -  & \textbf{95.18} & 93.29 & 90.22 & 94.0 & \textbf{94.5} \\
    \rowcolor{mycolor} \textbf{Point-CPR} & -  & MPM   & \textbf{2.7} & \textbf{1.9} & 93.80 & \textbf{93.46} & 88.72 & 93.6 & 94.1 \\
    \midrule
    \multicolumn{10}{c}{\textit{Multimodal Self-Supervised Learning}} \\
    \midrule
    % Joint-MAE~\cite{jointmae} & IJCAI 2023 & MMPM  & -     & -     & 90.94 & 88.86 & 86.07 & -     & 94.0 \\
    ACT~\cite{act}   & ICLR 2023 & MMPM  & 22.1  & 4.8   & 93.29 & 91.91 & 88.21 & 93.2  & 93.7 \\
    I2P-MAE~\cite{i2pmae} & CVPR 2023 & MMPM  & 15.3  & 3.6   & 94.15 & 91.57 & 90.11 & 93.7  & 94.1 \\
    TAP+PointMLP~\cite{tap} & ICCV 2023 & MMPM  & 12.6    & 31.4     & - & -  & 88.50 & 94.0    & - \\
    Recon~\cite{recon} & ICML 2023 & MMPM  & 44.3  & 5.3   & 95.18 & 93.29 & 90.63 & 94.1  & 94.5 \\
    \bottomrule
    \end{tabular}%
  }
  \caption{Classification accuracy on real-scanned (ScanObjectNN) and synthetic (ModelNet40) point clouds. In ScanObjectNN, we report the overall accuracy (\%) on three variants. In ModelNet40, we report the overall accuracy (\%) for both without and with voting. "\#Params" represents the model's parameters and GFLOPs refer to the model's floating point operations. CL, MPM, and MMPM respectively refer to pre-training strategies based on contrastive learning, single-modal masked point modeling, and multimodal masked point modeling.}
  \label{class}%
\end{table*}%

\subsection{Partial-aware Decoder}
\label{subsec:decoder}

To address the issue of limited representation caused by the position leakage of the center coordinates of invisible patches during point cloud reconstruction in vanilla MPM, we propose predicting the features of these invisible patches without relying on their center points as input. We introduce a partial-aware decoder to simultaneously reconstruct the center coordinates and semantic coordinates of invisible patches. Its structure is depicted in Figure \ref{framework}, consisting of $m$ stacked decoding layers. Each layer is composed of a partial-aware prediction module and a standard Transformer layer. We do not replace the original Transformer structure in the decoder, as the decoder is discarded during downstream fine-tuning.

\begin{figure}[t]
    \begin{center}
    \includegraphics[width=\linewidth]{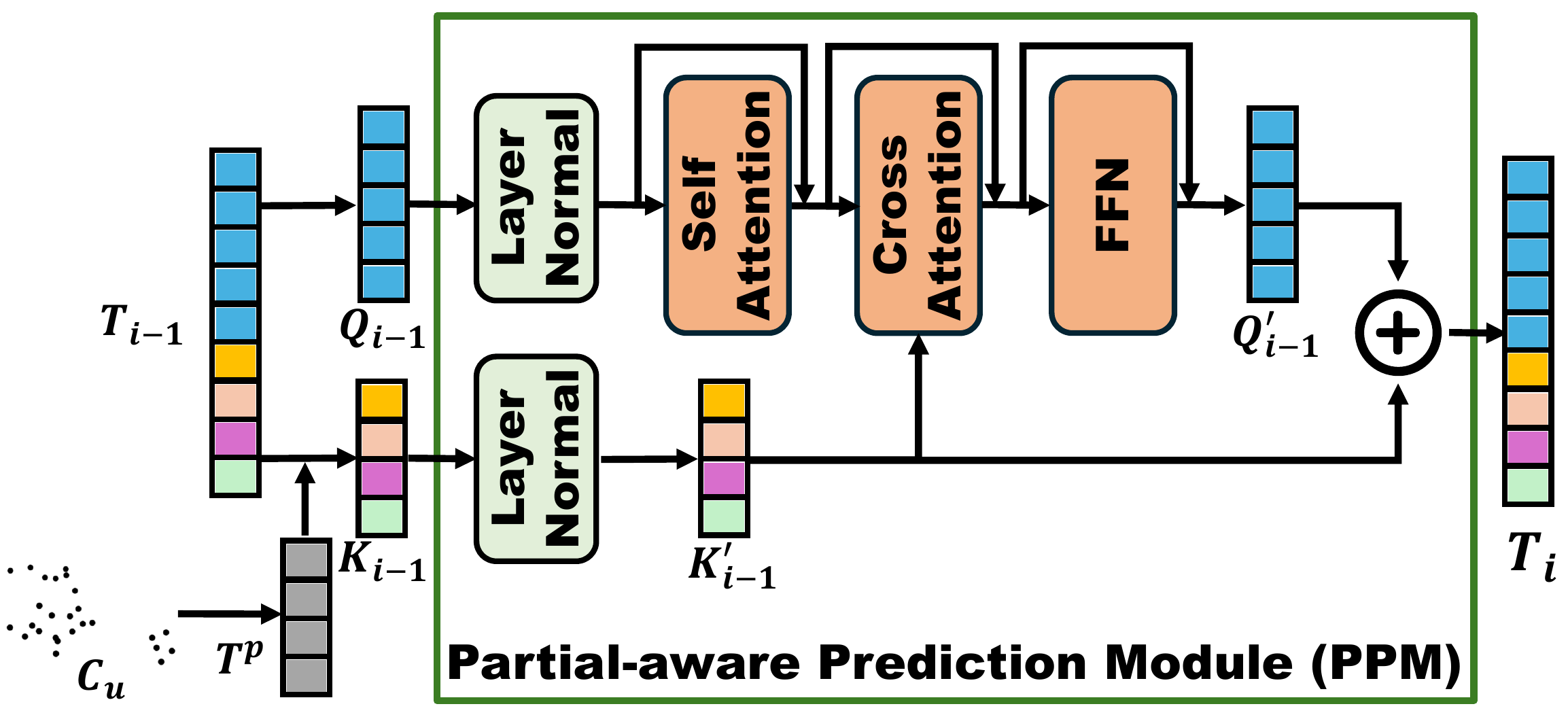}
    \caption{The structure of our partial-aware prediction module.
    }\label{decoder}
    \end{center}
\end{figure}

We rely on the partial-aware prediction module to predict the feature of invisible patches, and its structure is illustrated in Figure \ref{decoder}. For the $i$-th decoding layer, we first split the output $\bm {T_{i-1}}$ of the previous layer into masked query features $\bm {Q_{i-1}}$ representing invisible patches and features $\bm {K_{i-1}}$ representing visible patches. 
For the visible $\bm {K_{i-1}}$, we add position embedding $\bm {T^p}$ to it. 
Then, we apply layer normalization separately to $\bm {Q_{i-1}}$ and $\bm {K_{i-1}}$. Afterward, we apply self-attention to $\bm {Q_{i-1}}$, followed by adding the forward residual, to capture semantics across all query scales. Since $\bm {Q_{i-1}}$ represents all invisible patches, predicting the semantics of the invisible patches from the visible portion is crucial. Cross-attention, as a classical technique for capturing cross-domain semantic correlations, can effectively perceive the geometry of point clouds from visible regions and transfer this information to invisible queries. Therefore, we further employ cross-attention to predict invisible queries from visible patch tokens. 
Specifically, we perform cross-attention by using the $\bm {Q_{i-1}}$ as the query, and $\bm {K_{i-1}^{'}}$ as both key and value. Afterward, we use FFN for non-linear mapping to obtain a query $\bm {Q_{i-1}^{'}}$ that perceives the visible point cloud. 
Finally, we concatenate $\bm {Q_{i-1}^{'}}$  and $\bm {K_{i-1}^{'}}$ along the patch dimension to obtain the output of the Partial-aware Prediction Module (PPM), denoted as $\bm {T_{i}}$. We feed it into a standard Transformer layer for further decoding, as illustrated in Figure \ref{framework}.

\section{Experiments}

\subsection{Pre-training on ShapeNet}

For a fire comparison, we use ShapeNet~\cite{shapenet} as our pre-training dataset, encompassing over 50,000 distinct 3D models spanning 55 prevalent object categories. We extract 1024 points from each 3D model to serve as input for pre-training. The input point cloud is further divided into 64 point patches, with each patch containing 32 points.

\subsection{Fine-tuning on Downstream Tasks}
\label{subsec:exp}

We assess the performance of our pre-trained model by fine-tuning our models on various downstream tasks, including object classification, scene-level detection, part segmentation, and low-level completion tasks.

\subsubsection{Object Classification} 

We initially assess the overall classification accuracy of our pre-trained models on both real-scanned (ScanObjectNN~\cite{scanobjectnn}) and synthetic (ModelNet40~\cite{modelnet}) datasets. ScanObjectNN is a prevalent dataset consisting of approximately 15,000 real-world scanned point cloud samples from 15 categories. These objects represent indoor scenes and are often characterized by cluttered backgrounds and occlusions caused by other objects. ModelNet40 is a well-known synthetic point cloud dataset, comprising 12,311 meticulously crafted 3D CAD models distributed across 40 categories. We follow the practices of previous studies~\cite{act,recon,i2pmae}. For the ModelNet40 dataset, we sample 1024 points for each instance and report both the overall accuracy without voting and with voting. For the ScanObjectNN dataset, we sample 2048 points for each instance, employ data augmentation of simple rotations and report results without voting mechanisms.

As presented in Table \ref{class}, firstly, compared to recent MPM-based approaches, our Point-CPR achieves state-of-the-art performance in most datasets. Notably, our method demands minimal computational resources, \eg 2.7M parameters and 1.9 GFLOPs, which is significantly lower than other methods. Specifically, compared to the leading MPM approach, PointGPT-B~\cite{pointgpt}, our method utilizes only 2.2\% of its parameter count. The results of PointGPT-B are derived from the official repository and indicate the outcomes without post-pre-training. 
Secondly, our method surpasses the majority of multimodal MPM-based approaches (MMPM), ranking just below the leading Recon. This is still highly competitive as Recon~\cite{recon} benefits from the supplementary knowledge of image, and language modalities, while also requiring significantly more parameters than our method.

\subsubsection{Object Detection} We further assess the object detection performance of our pre-trained model on the more challenging scene-level point cloud dataset, ScanNet~\cite{scannet}, to evaluate our model's scene understanding capabilities. 
Following the previous pre-trained model~\cite{maskpoint,act}, we use 3DETR~\cite{3detr} as the baseline and only replace the Transformer-based encoder of 3DETR with our pre-trained compact encoder. Subsequently, the entire model is fine-tuned for object detection. In contrast to previous approaches~\cite{maskpoint,act,pimae}, which necessitate pre-train on large-scale scene-level point clouds like ScanNet, our approach directly utilizes models pre-trained on ShapeNet. This further emphasizes the generalizability of our pre-trained models to both object-level point clouds and scene-level point clouds.

% Table generated by Excel2LaTeX from sheet 'Sheet4'
\begin{table}[htbp]
  \centering
  \resizebox{\linewidth}{!}{
    \begin{tabular}{lccc}
    \toprule
     Methods          &  Pre-training     & $AP_{25}$  & $AP_{50}$ \\
    \midrule
    VoteNet~\cite{votenet}(\textit{baseline}) & \ding{56}  & 58.6  & 33.5 \\
    PointContrast~\cite{pointcontrast} & CL    & 58.5  & 38.0 \\
    STRL~\cite{strl}  & CL    & -     & 38.4 \\
    DepthContrast~\cite{depthcontrast} & CL    & 64.0  & 42.9 \\
    \midrule
    3DETR~\cite{3detr}(\textit{baseline}) & \ding{56}  & 62.1  & 37.9 \\
    Point-BERT~\cite{pointbert} & MPM   & 61.0  & 38.3 \\
    MaskPoint~\cite{maskpoint} & MPM   & 63.4  & 40.6 \\
    PiMAE~\cite{pimae} & MMPM  & 62.6  & 39.4 \\
    TAP~\cite{tap}   & MMPM  & 63.0  & 41.4 \\
    ACT~\cite{act}   & MMPM  & 63.8  & 42.1 \\
    % \rowcolor{mycolor} \textbf{Point-CPR} (from scratch) & \ding{56}   & \textbf{63.9}  & \textbf{46.9} \\
    \rowcolor{mycolor} \textbf{Point-CPR} & MPM   & \textbf{64.1}  & \textbf{48.7} \\
    \bottomrule
    \end{tabular}%
  }
  \caption{Object detection results on ScanNet. We adopt the average precision with 3D IoU thresholds of 0.25 ($AP_{25}$) and 0.5 ($AP_{50}$) for the evaluation metrics.}
  \label{dete}%
\end{table}%

Table \ref{dete} showcases our experimental findings, indicating a 6.6\%  improvement on $AP_{50}$ over the previous leading MPM method, ACT~\cite{act}. This demonstrates a notable superiority of our approach over prior contrastive learning-based and MPM-based methods. This superiority derives from two key aspects: firstly, our proposed compact encoder, owing to its more condensed local structural perception capability, notably surpasses existing Transformer-based encoders in scene analysis. Secondly, our introduced pre-trained model with partial-aware reconstruction also provides superior prior for object detection tasks.

\subsubsection{Part Segmentation}  We also assess the performance of Point-CPR in part segmentation using the ShapeNetPart dataset~\cite{shapenet}, comprising 16,881 samples across 16 categories. We utilize the same segmentation head after the pre-trained encoder as in previous works \cite{pointmae,m2ae} for fair comparison. The head only conducts simple upsampling for point tokens at different stages and concatenates them alone with the feature dimension as the output. As shown in Table \ref{seg}, our Point-CPR exhibits competitive performance among both existing CL-based and MPM-based methods and is slightly inferior to 2 MMPM-based methods. These results demonstrate that our approach exhibits superior performance in tasks such as part segmentation, which demands a more fine-grained understanding of point clouds.

\subsubsection{Point Cloud Completion} The previous pre-trained models required the centroid coordinates of the masked patches to serve as positional priors during the reconstruction, hence these models couldn't be directly applied to real point cloud reconstruction tasks. Our pre-trained model benefits from the proposed partial-aware reconstruction mechanism, enabling our decoder entirely independent of any unknown point cloud priors. Consequently, it can be directly applied to low-level reconstruction tasks, such as point cloud completion. Therefore, we first evaluate the performance of our pre-trained model on the task of point cloud completion and compare it with the state-of-the-art point cloud completion methods.

\begin{table}[t]
  \centering
  \resizebox{\linewidth}{!}{
    \begin{tabular}{lccc}
    \toprule
    Methods & Pre-training & $\mathrm{mIoU}_{c}$ & $\mathrm{mIoU}_{I}$ \\
    \midrule
    \multicolumn{4}{c}{\textit{Supervised Learning Only}} \\
    \midrule
    PointNet~\cite{pointnet}  &  \ding{56} & 80.4  & 83.7 \\
    PointNet++~\cite{pointnet++} &  \ding{56} & 81.9  & 85.1 \\
    DGCNN~\cite{dgcnn}  &  \ding{56} & 82.3  & 85.2 \\
    PointMLP~\cite{pointmlp} & \ding{56} & 84.6  & 86.1 \\
    \midrule
    \multicolumn{4}{c}{\textit{Single-Modal Self-Supervised Learning}} \\
    \midrule
    Transformer~\cite{attention} & MPM & 83.4  & 84.7 \\
    Transformer-OcCo~\cite{wang2021unsupervised} & CL & 83.4  & 85.1 \\
    PointContrast~\cite{pointcontrast} & CL & -  & 85.1 \\
    CrossPoint~\cite{crosspoint} & CL & -  & 85.5 \\
    IDPT~\cite{idpt} & MPM & 83.8  & 85.9 \\
    Point-BERT~\cite{pointbert} & MPM & 84.1  & 85.6 \\
    MaskPoint~\cite{maskpoint} & MPM & 84.4  & 86.0 \\
    Point-MAE~\cite{pointmae} & MPM & 84.2  & 86.1 \\
    PointGPT-S~\cite{pointgpt} & MPM & 84.1  & 86.2 \\
    PointGPT-B~\cite{pointgpt} & MPM & 84.5  & 86.4 \\
    PointFEMAE~\cite{femae} & MPM & 84.9  & 86.3 \\
    Point-M2AE~\cite{m2ae} & MPM & 84.9  & \textbf{86.5} \\
    \rowcolor{mycolor} \textbf{Point-CPR} & MPM & \textbf{85.1} & \textbf{86.5} \\
    \midrule
    \multicolumn{4}{c}{\textit{Multimodal Self-Supervised Learning}} \\
    \midrule
    ACT~\cite{act} & MMPM   & 84.7  & 86.1 \\
    Joint-MAE~\cite{jointmae} & MMPM   & 85.4  & 86.3 \\
    Recon~\cite{recon} & MMPM & 84.8  & 86.4 \\
    I2P-MAE~\cite{i2pmae} & MMPM   & 85.2  & 86.8 \\
    TAP+PointMLP~\cite{tap} & MMPM   & 85.2  & 86.9 \\
    \bottomrule
    \end{tabular}%
  }
  \caption{Part segmentation results on the ShapeNetPart. The mean IoU across all categories, i.e., $\mathrm{mIoU}_{c}$ (\%), and the mean IoU across all instances, i.e., $\mathrm{mIoU}_{I}$ (\%) are reported.}
  \label{seg}%
\end{table}%

We evaluate the transferability of our pre-trained model to low-level tasks by fine-tuning it on the classic point cloud completion dataset, PCN \cite{pcn}. The PCN dataset is created from the ShapeNet dataset, including eight categories with a total of 30974 CAD models. We followed the data processing methods established in previous works~\cite{pointr,proxyformer} and used the $l_1$ Chamfer Distance~\cite{cdloss} to evaluate the results. As shown in Table \ref{comp}, our approach exhibits substantial improvements in aircraft, cabinet, and car completion compared to the previous leading method ProxyFormer~\cite{proxyformer} and achieves the lowest average Chamfer Distance, highlighting the significant advantage of our partial-aware reconstruction for point cloud pre-training.

\begin{table}[htbp]
  \centering
  \resizebox{\linewidth}{!}{
    \begin{tabular}{lccccccccc}
    \toprule
    \multirow{2}[4]{*}{Methods} & \multicolumn{9}{c}{Chamfer Distance ($10^{-3}$ \textcolor{blue}{($\downarrow$)})} \\
\cmidrule{2-10}          & \textbf{Ave}   & Air.   & Cab.   & Car   & Cha.   & Lam.   & Sof.   & Tab.   & Ves. \\
    \midrule
    FoldingNet~\cite{foldingnet} & 14.31 & 9.49  & 15.80 & 12.61 & 15.55 & 16.41 & 15.97 & 13.65 & 14.99 \\
    AtlasNet~\cite{atlasnet} & 10.85 & 6.37  & 11.94 & 10.10 & 12.06 & 12.37 & 12.99 & 10.33 & 10.61 \\
    PCN~\cite{pcn}   & 9.64  & 5.50  & 22.70 & 10.63 & 8.70  & 11.00 & 11.34 & 11.68 & 8.59 \\
    CRN~\cite{crn}   & 8.51  & 4.79  & 9.97  & 8.31  & 9.49  & 8.94  & 10.69 & 7.81  & 8.05 \\
    PMP-Net~\cite{pmpnet} & 8.73  & 5.65  & 11.24 & 9.64  & 9.51  & 6.95  & 10.83 & 8.72  & 7.25 \\
    PoinTr~\cite{pointr} & 8.38  & 4.75  & 10.47 & 8.68  & 9.39  & 7.75  & 10.93 & 7.78  & 7.29 \\
    SnowflakeNet~\cite{snowflakenet} & 7.21  & 4.29  & 9.16  & 8.08  & 7.89  & 6.07  & 9.23  & 6.55  & 6.40 \\
    ProxyFormer~\cite{proxyformer} & 6.77  & 4.01  & 9.01  & 7.88  & \textbf{7.11} & \textbf{5.35} & 8.77  & \textbf{6.03} & \textbf{5.98} \\
    % \rowcolor{mycolor} \textbf{Point-CPR} (from scratch) &       &       &       &       &       &       &       &       &  \\
    \rowcolor{mycolor} \textbf{Point-CPR} (Ours)& \textbf{6.75} & \textbf{3.75} & \textbf{8.81} & \textbf{7.46} & 7.35  & 5.71  & \textbf{8.69} & 6.27  & \textbf{5.98} \\
    \bottomrule
    \end{tabular}%
  }
  \caption{Quantitative comparison of point cloud completion task on PCN. Point resolutions for the output and GT are 16384. For Chamfer Distance, lower is better.}
  \label{comp}%
\end{table}%

\subsection{Ablation Study}

\subsubsection{Effects of Position Leakage in Decoder}

\begin{table}[htbp]
  \centering
  \resizebox{\linewidth}{!}{
    \begin{tabular}{ccllll}
    \toprule
    \multirow{2}[4]{*}{Index} &\multirow{2}[4]{*}{Pre-training} & \multicolumn{2}{c}{ScanObjectNN Classification} & \multicolumn{2}{c}{ScanNetV2 Detection} \\
    \cmidrule(lr){3-4}\cmidrule(lr){5-6}   &     & OBJ-ONLY & PB-T50-RS & $AP_{25}$  & $AP_{50}$ \\
    \midrule
    A & \ding{56}   & 90.71 & 87.47 & 63.8      & 46.4 \\
    B &Vanilla MPM w/o pos & 91.22\textcolor{blue}{($\uparrow$ 0.51)} & 87.86\textcolor{blue}{($\uparrow$ 0.39)} & 63.6\textcolor{gray}{($\downarrow$ 0.2)}      & 47.0\textcolor{blue}{($\uparrow$ 0.6)} \\
    C &Vanilla MPM  & 92.43\textcolor{blue}{($\uparrow$ 1.72)} & 88.13\textcolor{blue}{($\uparrow$ 0.66)} & 63.9\textcolor{blue}{($\uparrow$ 0.1)}      & 47.3\textcolor{blue}{($\uparrow$ 0.9)} \\
    D &Partial-aware (Ours)    & \textbf{93.46}\textcolor{blue}{($\uparrow$ 2.75)} & \textbf{88.72}\textcolor{blue}{($\uparrow$ 1.25)} & \textbf{64.1}\textcolor{blue}{($\uparrow$ 0.3)} & \textbf{48.7}\textcolor{blue}{($\uparrow$ 2.3)} \\
    \bottomrule
    \end{tabular}%
  }
  \caption{Comparison of the impact of partial-aware reconstruction strategy and vanilla MPM reconstruction strategy on classification and detection tasks. 
  }
  \label{abl1}%
\end{table}%

To illustrate the issue of position leakage in the decoding phase of the previous MPM method, we conducted four sets of experiments using the compact encoder settings in classification and detection tasks. A corresponds to training from scratch, B uses the vanilla MPM strategy but does not include position encoding for masked patches in the decoder, C employs the vanilla MPM strategy, and D represents our partial-aware reconstruction strategy.

As shown in Figure \ref{abl1}, comparing A with B, C, and D reveals that the pre-training strategy indeed enhances representational capacity. The comparison between B and C indicates the crucial role of position embedding for masked patches in vanilla MPM pre-training, as it simplifies the reconstruction task, allowing the decoder to learn shortcuts between the center coordinates of masked patches and the relative coordinates of masked patches. However, this also further limits the potential for representation learning. Comparing C and D, our method does not use the center coordinates of masked patches as input, making the reconstruction task more challenging, but our partial-aware reconstruction mechanism can learn the unknown from the known, further unleashing the potential for representation learning. The comparison between B and D further illustrates that under the same decoder input, our partial-aware reconstruction greatly unleashes the potential for representation learning. These results strongly demonstrate the effectiveness of our proposed partial-aware reconstruction.

\subsubsection{Effects of Compact Encoder}
\label{abl}

% Table generated by Excel2LaTeX from sheet 'Abl'
\begin{table}[htbp]
  \centering
  \resizebox{\linewidth}{!}{
    \begin{tabular}{lllll}
    \toprule
    \multirow{2}[4]{*}{Encoder} & \multicolumn{2}{l}{ScanObjectNN Classification} & \multicolumn{2}{l}{ScanNetV2 Detection} \\
    \cmidrule(lr){2-3}\cmidrule(lr){4-5}          & \# Params.(M) & Overall Acc. & $AP_{25}$  & $AP_{50}$ \\
    \midrule
    Transformer & 22.1 & 86.92 & 60.5  & 40.6 \\
    Compact (Ours)  & \textbf{2.7}\textcolor{blue}{($\downarrow$ 19.4)} & \textbf{87.47}\textcolor{blue}{($\uparrow$ 0.55)} & \textbf{63.8}\textcolor{blue}{($\uparrow$ 3.3)} & \textbf{46.4}\textcolor{blue}{($\uparrow$ 5.8)} \\
    \bottomrule
    \end{tabular}%
  }
  \caption{Effects of Different Encoder. We compare the performance of the Transformer-based encoder and our compact encoder on classification and detection tasks, all experiments are training from scratch.}
  \label{enc}%
\end{table}%

We explore the performance of our compact encoder by comparing it with a Transformer-based encoder in classification and detection tasks. 
Table \ref{enc} illustrates the classification performance on the real scanned point cloud dataset ScanObjectNN and the detection performance on the ScanNetV2 dataset for two different networks without any pre-training. Our compact encoder outperforms the Transformer-based structure significantly, particularly in detection tasks, with an improvement of up to 5.8\% in the $AP_{50}$ metric. 

This substantial enhancement is attributed to the compact encoder's focus on local neighborhood information, crucial for point cloud analysis, especially in large-scale scene-level point clouds. Another contributing factor is the utilization of fewer network parameters; for instance, in classification, we only require 2.7M parameters compared to the Transformer-based model's 22.1M, which better mitigates network overfitting in existing limited-size point cloud datasets. We provide a detailed explanation of this overfitting phenomenon in the supplementary material. Please refer to the supplementary material for more details.

\section{Conclusion}

In this paper, we propose Point-CPR, a masking point modeling pre-training framework to solve the limitations of the existing MPM-based methods in practical applications. We first propose partial-aware reconstruction, replacing existing position-embedded masked tokens with random masked queries and leveraging a partial-aware predict module in the decoder to predict it. This process resolves the limited 3D representations resulting from position leakage of masked patches in previous methods. Secondly, we introduce a compact encoder composed solely of local aggregations and MLPs, replacing the previous Transformer-based encoder to mitigate demands on practical device resources caused by oversized models. Finally, extensive experiments validate the superiority of our proposed approach in both efficiency and performance.

%% The file named.bst is a bibliography style file for BibTeX 0.99c
\bibliographystyle{named}
\bibliography{ijcai24}

\end{document}